\documentclass{article}

\usepackage{microtype}
\usepackage{graphicx}
\usepackage{booktabs} 
\usepackage{float}

\usepackage{hyperref}

\usepackage[accepted]{icml2020}

\usepackage[utf8]{inputenc} 
\usepackage[T1]{fontenc}    
\usepackage{url}            
\usepackage{booktabs}       
\usepackage{amsfonts}       
\usepackage{nicefrac}       
\usepackage{amsmath}
\usepackage{amssymb}
\usepackage{verbatim}
\usepackage{tikz}
\usetikzlibrary{shapes.geometric}
\usepackage{float}
\usepackage{subfig}
\usepackage{xstring}
\usepackage{tikz}
\usepackage{pifont}
\usetikzlibrary{automata,calc,backgrounds,arrows,positioning, shapes.misc,quotes,arrows.meta, decorations,decorations.markings}
\usepackage{color}
\usepackage{pgfplots}
\usepackage{mathtools}
\usepackage{array}
\usepackage{multirow}
\usepackage{rotating}
\usepackage{wrapfig}
\usepackage{caption}

\newcommand{\vc}[1]{{\pmb{#1}}}
\newcommand{\x}{{\pmb{x}}}

\newcommand{\z}{{\pmb{z}}}

\newcommand{\h}{{\pmb{h}}}

\renewcommand{\b}{{\pmb{b}}}

\renewcommand{\l}{{\pmb{l}}}

\newcommand{\W}{{\pmb{W}}}
\newcommand{\btheta}{{\pmb{\theta}}}
\newcommand{\bzeta}{{\pmb{\zeta}}}
\newcommand{\bepsilon}{{\pmb{\epsilon}}}
\newcommand{\bphi}{{\pmb{\phi}}}
\newcommand{\blambda}{{\pmb{\Lambda}}}
\newcommand{\bmu}{{\pmb{\mu}}}

\renewcommand{\L}{{\mathcal{L}}}
\newcommand{\K}{{\mathcal{K}}}
\newcommand{\U}{{\mathcal{U}}}
\newcommand{\E}{{\mathbb{E}}}

\def\KL{\text{KL}}

\newcommand{\be}{\begin{equation}}
\newcommand{\ee}{\end{equation}}
\newcommand{\ba}{\begin{align}}
\newcommand{\ea}{\end{align}}

\renewcommand{\algorithmiccomment}[1]{$\rhd$ #1}

\icmltitlerunning{Undirected Graphical Models as Approximate Posteriors}

\begin{document}

\twocolumn[
\icmltitle{Undirected Graphical Models as Approximate Posteriors}



\icmlsetsymbol{equal}{*}

\begin{icmlauthorlist}
\icmlauthor{Arash Vahdat}{nvidia}
\icmlauthor{Evgeny Andriyash}{sanctuary}
\icmlauthor{William G. Macready}{sanctuary}
\end{icmlauthorlist}

\icmlaffiliation{nvidia}{NVIDIA, USA}
\icmlaffiliation{sanctuary}{Sanctuary AI, Canada, Work done at D-Wave Systems}

\icmlcorrespondingauthor{Arash Vahdat}{avahdat@nvidia.com}

\icmlkeywords{Machine Learning, ICML, Undirected Graphical Models, UGM, Stochastic Optimization, Variational Autoencoders, VAE, Approximate Posterior}

\vskip 0.3in
]



\printAffiliationsAndNotice{}  

\begin{abstract}
The representation of the approximate posterior is a critical aspect of effective variational autoencoders (VAEs). Poor choices for the approximate posterior have a detrimental impact on the generative performance of VAEs due to the mismatch with the true posterior. We extend the class of posterior models that may be learned by using undirected graphical models. We develop an efficient method to train undirected approximate posteriors by showing that the gradient of the training objective with respect to the parameters of the undirected posterior can be computed by backpropagation through Markov chain Monte Carlo updates. We apply these gradient estimators for training discrete VAEs with Boltzmann machines as approximate posteriors and demonstrate that undirected models outperform previous results obtained using directed graphical models. Our implementation is available \href{https://github.com/QuadrantAI/dvaess}{here}.
\end{abstract}

\section{Introduction}
Training of likelihood-based deep generative models has advanced rapidly in recent years. These advances have been enabled by amortized inference~\citep{hinton1995wake, mnih2014neural, gregorICML14}, which scales up training of variational models, the reparameterization trick~\citep{kingma2014vae, rezende2014stochastic}, which provides low-variance gradient estimates, and increasingly expressive neural networks. Combinations of these techniques have resulted in many different forms of variational autoencoders (VAEs)~\citep{kingma2016improved, chen2016variational}.

It is also widely recognized that flexible approximate posterior distributions improve the generative quality of variational autoencoders (VAEs) by more faithfully modeling true posterior distributions.   
More accurate approximate posterior models have been realized by reducing the gap between true and approximate posteriors with tighter bounds~\citep{burda2015importance}, and using auto-regressive architectures~\citep{gregorICML14, gregor2015draw} or flow models~\citep{rezendeICML15}.

A complementary (but unexplored) direction for further improving the richness of approximate posteriors is available through the use of undirected graphical models (UGMs). In this approach, the approximate posterior over latent variables is formed using UGMs whose parameters are inferred for each observed sample in an amortized fashion similar to VAEs. This is compelling as UGMs can succinctly capture complex relationships between variables. However, there are three challenges when training UGMs as approximate posteriors: i) There is no known low-variance path-wise gradient estimator (like the reparameterization trick) for learning parameters of UGMs. ii) Sampling from general UGMs is intractable and approximate sampling is computationally expensive.  iii) Evaluating the probability of a sample under a UGM can require an intractable partition function. These two latter costs are particularly acute when UGMs are used as posterior approximators because the number of UGMs grows with the size of the dataset.

However, we note that posterior\footnote{We may use ``posterior'' when referring to ``approximate posterior'' for brevity. To disambiguate, we always refer to ``true posterior'' explicitly.} UGMs conditioned on observed data points are often simple as there are usually only a small number of modes in latent space explaining an observation. We expect then, that sampling and partition function estimation (challenges ii and iii) can be solved efficiently using parallelizable Markov chain Monte Carlo (MCMC) methods. In fact, we observe that UGM posteriors trained in a VAE model tend to be unimodal but with a mode structure that is not necessarily well-captured by mean-field posteriors. Nevertheless, in this case, MCMC-based sampling methods quickly equilibrate.

To address challenge i) we estimate the gradient by reparameterized sampling in the MCMC updates. Although an infinite number of MCMC updates are theoretically required to obtain an unbiased gradient estimator, we observe that a single MCMC update provides a low-variance but biased gradient estimation that is sufficient for training UGMs.

Binary UGMs in the form of Boltzmann machines~\citep{ackley1985learning} have recently been shown to be effective as priors for VAEs allowing for discrete versions of VAEs~\citep{rolfe2016discrete, Vahdat2018DVAE++, vahdat2018dvaes}. However, previous work in this area has relied on directed graphical models (DGMs) for posterior approximation. 
In this paper, we replace the DGM posteriors of discrete VAEs (DVAEs) with Boltzmann machine UGMs and show that these posteriors provide a generative performance comparable to or better than previous DVAEs. We denote this model as DVAE\#\#, where \#\# indicates the use of UGMs in the prior and the posterior.

We begin by summarizing related work on developing expressive posterior models including models trained nonvariationally. Nonvariational models employ MCMC to directly sample posteriors and differ from our variational approach which uses MCMC to sample from an amortized undirected posterior. Sec.~\ref{sec:background} provides the necessary background on variational learning and MCMC to allow for the development in Sec.~\ref{sec:learning} of a gradient estimator for undirected posteriors. We provide examples for both Gaussian (Sec.~\ref{sec:toy}) and Boltzmann machine (Sec.~\ref{sec:rbm}) UGMs. Experimental results are provided in Sec.~\ref{sec:experiments} on VAEs (Sec.~\ref{sec:vae-expt}), importance-weighted VAEs (Sec.~\ref{sec:iwvae-expt}), and structured prediction (Sec.~\ref{sec:sp-expt}), where we observe consistent improvement using UGMs. We conclude in Sec.~\ref{sec:conclusion} with a list of future work.

\subsection{Related Work}
\textbf{Inference gap reduction in VAEs:} Previous work on reducing the gap between true and approximate posteriors can be grouped into three categories: i) Training objectives that replace the variational bound with tighter bounds~\citep{burda2015importance, li2016renyi, bornschein2016bidirectional}. ii) Autoregressive models~\citep{hochreiter1997long, graves2013generating} that use DGMs to form more flexible distributions~\citep{gregorICML14, gulrajani2016pixelvae, van2016pixel, Salimans2017PixeCNN, chen2018pixelsnail}. iii) Flow-based models~\citep{rezendeICML15} that map data to a latent space using a class of invertible functions~\citep{kingma2016improved, kingma2018glow, dinh2014nice, dinh2016density}. To the best of our knowledge, UGMs have not been used as approximate posteriors.


\textbf{Gradient estimation in latent variable models:} REINFORCE~\citep{williams1992simple} is the most generic approach for computing the gradient of the approximate posteriors but suffers from high-variance and must be augmented by variance reduction techniques. For many continuous latent-variable models the reparameterization trick~\citep{kingma2014vae, rezende2014stochastic} provides lower-variance gradient estimates.
Reparameterization does not apply to discrete latent variables and recent methods for discrete variables have focused on REINFORCE with control variates~\citep{mnih2014neural, gu2015muprop, mnih2016variational, tucker2017rebar, grathwohl2017backpropagation} or continuous relaxations~\citep{maddison2016concrete, jang2016categorical, rolfe2016discrete, Vahdat2018DVAE++, vahdat2018dvaes}. See~\citep{andriyash2018improved} for a review.

\textbf{Variational Inference and MCMC:} A common alternative to variational inference is to use MCMC to sample directly from true posteriors in generative models~\citep{welling2011bayesian, salimans2015markov, wolf2016variational,hoffman2017learning, li2017approximate, caterini2018hamiltonian}. However, this approach is often computationally intensive, as it requires computing the prior and decoder distributions (often implemented by neural networks) many times at each each parameter update for each training data point in a batch. Moreover, evaluating these models is more challenging, as there is no amortized approximating posterior for importance sampling~\cite{burda2015importance}. Our method differs from these techniques, as we use MCMC to sample from an amortized approximate posterior represented by UGMs.

\section{Background} \label{sec:background}
In this section we provide background for the topics discussed in this paper.

\textbf{Undirected graphical models:}
A UGM represents the joint probability distribution for a set of random variables $\z$ as $q(\z) = \exp(-E_{\bphi}(\z))/Z_\bphi$, where $E_{\bphi}$ is a $\bphi$-parameterized energy function defined by an undirected graphical model, and $Z_{\bphi} = \int d\z \exp\bigl(-E_{\bphi}(\z)\bigr)$ is the partition function. UGMs over binary variables $\z$ with bipartite quadratic energy functions $E_{\bphi}(\z_1, \z_2) = \b_1^T\z_1 + \b_2^T\z_2 + \z_1^T \W \z_2$ are called restricted Boltzmann machines (RBMs). Parameters $\bphi=\{\b_1, \b_2, \W\}$ encode linear biases and pairwise interactions. RBMs permit parallel Gibbs sampling updates as $q_\bphi(\z_1|\z_2)$ and $q_\bphi(\z_2|\z_1)$ are factorial in the components of $\z_1$ and $\z_2$.

\textbf{Variational autoencoders:} A VAE is a generative model factored as $p(\x, \z) = p(\z)p(\x | \z)$, where $p(\z)$ is a prior distribution over latent variables $\z$ and $p(\x | \z)$ is a probabilistic decoder representing the conditional distribution over data variables $\x$ given $\z$. VAEs are trained by maximizing a variational lower bound (ELBO) on $\log p(\x)$:
\be \label{eq:vae}
\L = \E_{q(\z|\x)} \biggl[ \log \frac{p(\x, \z)}{q(\z|\x)} \biggr] \leq \log p(\x),
\ee
where $q(\z|\x)$ is a probabilistic encoder that approximates the posterior over latent variables given a data point. 
A tighter importance-weighted bound is obtained with
\be \label{eq:iw}
\L_{IW} = \E_{\z_{1:K}} \left[\log \left( \frac{1}{K} \sum_{i=1}^{K} \frac{p(\z_i, \x)}{q(\z_i|\x)} \right) \right] \leq \log p(\x),
\ee
where $\z_{1:K}\sim \prod_i q(\z_i|\x)$.  For continuous latent variables, these bounds are optimized using the reparameterization trick~\citep{kingma2014vae, rezende2014stochastic}. However, as noted above, continuous relaxations are commonly employed with discrete latent variables.

\textbf{MCMC:} MCMC methods are used to draw approximate samples from a target distribution $q(\z)$. Each MCMC method is characterized by a transition kernel $\K(\z|\z')$  designed so that
\be \label{eq:fixed_point}
q(\z) = \int d\z' q(\z') \K(\z|\z') \quad \quad \forall \z.
\ee
Samples from the fixed point $q(\z)$ are found by sampling from an initial distribution $q_0(\z)$ and iteratively updating the samples by drawing conditional samples using $\z_t|\z_{t-1} \sim \K(\z_t|\z_{t-1})$. Denoting the distribution of the samples after $t$ iterations by $q_t=\K^t q_0$, the theory of MCMC shows that under some regularity conditions $q_t \rightarrow q$ as $t \rightarrow \infty$.
In practice, $\K(\z|\z')$ is usually chosen to satisfy the detailed balance condition:
\be \label{eq:detailed}
q(\z) \K(\z'|\z) = q(\z') \K(\z|\z') \quad \quad \forall \z,\z'
\ee
which implies Eq.~\eqref{eq:fixed_point}. $\K^t$ itself is a valid transition kernel and satisfies Eq.~\eqref{eq:fixed_point} and Eq.~\eqref{eq:detailed}.

Gibbs sampling is a particularly efficient MCMC method when variables $\z=[\z_1,\z_2]$ are partitioned as in an RBM. The transition kernel in this case may be written as
\be \label{eq:gibbs_kernel}
\K_{\text{\tiny Gibbs}}(\z_1, \z_2|\z_1', \z_2') = q(\z_2|\z_1)q(\z_1|\z_2').
\ee
where all partitioned variables in $\z_1$ or $\z_2$ may be updated in parallel. Gibbs sampling is a special MCMC method that does not satisfy the detailed balance condition. However, it does admit a reverse kernel in the form: 
\be \label{eq:rev_kernel}
\K_{\text{\tiny Reverse}}(\z_1', \z_2'|\z_1, \z_2) = q(\z_1'|\z_2')q(\z_2'|\z_1),
\ee
which samples from the variables in a reversed order.
Thus, for Gibbs sampling on bipartite UGMs, we have: 
\be \label{eq:gibbs_detail}
q(\z')\K_{\text{\tiny Gibbs}}(\z|\z') = q(\z)\K_{\text{\tiny Reverse}}(\z'|\z)  \quad \quad \forall \z,\z',
\ee
where $\z=[\z_1,\z_2]$  and $\z'=[\z_1',\z_2']$.

\section{Undirected Approximate Posteriors} \label{sec:learning}
In this section, we propose a gradient estimator for generic UGM approximate posteriors and discuss how the estimator is applied to RBM-based approximate posteriors.

\tikzset{
  strike through/.style={
    postaction=decorate,
    decoration={
      markings,
      mark=at position 0.5 with {
        \draw[-, double] (-2pt,-2pt) -- (3pt, 3pt);
      }
    }
  }
}

\begin{figure}
    \centering
    \begin{tikzpicture}[->,>=stealth',shorten >=1pt,auto,node distance=2.5cm, thick, scale=1.]
    \tikzstyle{every state}=[fill=white,draw=none,text=black, transform shape, minimum size=0.7cm]
    \node[state]   	             (x)        {$q_{\bphi}$};
    \node[state, right of=x] 	(zp)       {$\z'$};
    \node[state, right of=zp] 	 (z)       {$\z$};
    \node[state, right of=z]   	 (f)       {$\L$};
    \draw[>=latex,->,draw=blue] ([yshift= 3pt] x.east) -- ([yshift= 3pt] zp.west) node[pos=0.5]{$\z' \sim q_{\bphi}(\z')$};
    \draw[>=latex,<-,draw=red, strike through] ([yshift= -3pt] x.east) -- ([yshift= -3pt] zp.west) node[pos=0.5]{};
    \draw[>=latex,->,draw=blue] ([yshift= 3pt] zp.east) -- ([yshift= 3pt] z.west) node[pos=0.5]{$\z \sim K_{\bphi}(\z|\z')$};
    \draw[>=latex,<-,draw=red] ([yshift=-3pt] zp.east) -- ([yshift=-3pt] z.west) node[pos=0.5, below]{$\partial_\bphi \z(\bepsilon, \bphi, \z')$};
    \draw[>=latex,->,draw=blue] ([yshift= 3pt] z.east) -- ([yshift= 3pt] f.west) node[pos=0.5]{$f(\z)$};
    \draw[>=latex,<-,draw=red] ([yshift=-3pt] z.east) -- ([yshift=-3pt] f.west) node[pos=0.5, below]{$\partial_\z f(\z)$};
  \end{tikzpicture}  
    \caption{To estimate the gradient of $\L=\E_{q_\bphi(\z)}[f(\z)]$ w.r.t the parameters of the UGM posterior ($\bphi$), we first sample from the approximating posterior ($\z' \sim q_{\bphi}(\z')$), then, we apply an MCMC update via reparameterized sampling ($\z \sim K_{\bphi}(\z|\z')$). Finally, we evaluate the function on the samples $(f(\z))$. The gradient is computed automatically by {\color{red}backpropagating} through the reparameterized samples while ignoring the dependency of $\z'$ on $\bphi$.}
    \label{fig:big_picture}
\end{figure}

The training objective for a probabilistic encoder network with $q_\bphi(\z)$ being a UGM can be written as $\max_\bphi \E_{q_\bphi(\z)}[f(\z)]$,
where $f$ is a differentiable function of $\z$. For simplicity of exposition, we assume that $f$ does not depend on $\bphi$.\footnote{If $f$ does depend on $\bphi$, then $\E_{q_\bphi(\z)}[{\partial_\bphi} f(\z)]$ is approximated with Monte Carlo sampling.} Using the fixed point equation in Eq.~\eqref{eq:fixed_point}, we have:
\be \label{eq:gen_obj2}
\E_{q_\bphi(\z)}[f(\z)] = \E_{q_\bphi(\z')} \left[\E_{\K^t_{\bphi}(\z|\z')} [f(\z)] \right],
\ee
where the right hand side implies sampling $\z' \sim q_\bphi(\z')$ and applying MCMC updates $t$ times. To maximize Eq.~\eqref{eq:gen_obj2}, we require its gradient:
\begin{align} \label{eq:grad_all} 
 \partial_\bphi \E_{q_\bphi(\z)}[f(\z)] = & 
 \underbrace{ \int d\z' \,{\partial_\bphi} q_\bphi(\z') \E_{\K^t_{\bphi}(\z|\z')} [f(\z)]}_{I} \nonumber \\
+ & \underbrace{ \E_{q_\bphi(\z')} \left[ {\partial_\bphi} \E_{\K^t_{\bphi}(\z|\z')} [f(\z)] \right]}_{II}. 
\end{align}
The term marked as $I$ is  written as: 
\begin{align}
I &= \E_{q_\bphi(\z')\K^t_{\bphi}(\z|\z')} \bigl[ f(\z) \partial_\bphi \log q_\bphi(\z') \bigr] \label{eq:reinforce} \\
& = \E_{q_\bphi(\z)\K^t_{\bphi}(\z'|\z)} \bigl[f(\z) \partial_\bphi \log q_\bphi(\z') \bigr] \label{eq:reinforce2} \\
& = \E_{q_\bphi(\z)} \Bigl[ f(\z) \E_{\K^t_{\bphi}(\z'|\z)} \bigl[{\partial_\bphi} \log q_\bphi(\z') \bigr] \Bigr], \label{eq:score_func}
\end{align}
where we have used detailed balance to get Eq.~\eqref{eq:reinforce2} from Eq.~\eqref{eq:reinforce}. The form of Eq.~\eqref{eq:score_func} makes it clear that $I$ goes to zero as $t \rightarrow \infty$, because $\K^t_\bphi(\z'|\z) \rightarrow q_\bphi(\z')$ and $\E_{q_\bphi(\z')} \bigl[{\partial_\bphi} \log q_\bphi(\z') \bigr] = 0$. However, similar to REINFORCE, Monte-Carlo estimate of $I$ will have high variance due to the presence of $q_\bphi(\z')$ in the denominator\footnote{Technically, Gibbs sampling does not satisfy detailed balance. Instead, we can use the identity in Eq.~\eqref{eq:gibbs_detail} to derive the same argument for Gibbs sampling. In this case, the transition kernel in Eq.~\eqref{eq:reinforce2} and Eq.~\eqref{eq:score_func} will be the reverse kernel.}.

The expectation in Eq.~\eqref{eq:grad_all} marked as $II$ involves the gradient of an expectation with respect to MCMC transition kernel. The reparameterization trick provides a low-variance gradient estimator for this term, and as $t \to \infty$, the $II$ contribution approaches the full gradient because $I \to 0$.

Our key insight in this paper is that, given the high variance of $I$, it is beneficial to drop this term from the gradient (Eq.~\eqref{eq:grad_all}) and use the biased but lower-variance estimate $II$. The choice of $t$ trades increased computational complexity for decreased bias. We observe that increasing $t$ has little effect on the optimization performance. Therefore, $t=1$ is a good choice giving the smallest computational complexity.
Consequently, we use the approximation:
\begin{align}\label{eq:grad_biased}
\partial_{\bphi} \E_{q_\bphi(\z)}[f(\z)] &\approx \E_{q_\bphi(\z')} \Bigl[ {\partial_\bphi} \E_{\K_{\bphi}(\z|\z')} \bigl[f(\z)\bigr] \Bigr] \\
&= \E_{q_\bphi(\z')} \E_{\bepsilon \sim p(\bepsilon)} \bigl[ (\partial_\bphi \z) \partial_\z f(\z) \bigr], \nonumber
\end{align}
where $\z = \z(\bepsilon, \bphi, \z')$ is a reparameterized sample from $\K_{\bphi}(\z|\z')$ where $\bepsilon$ is a sample from a base distribution. Our gradient estimator is illustrated in Fig.~\ref{fig:big_picture}. The extension to $t>1$ Gibbs updates is straightforward.

In principle, MCMC methods such as Metropolis-Hastings or
Hamiltonian Monte Carlo (HMC) can be used as the transition kernel in Eq.~\eqref{eq:grad_biased}. However, unbiased reparameterized sampling for these methods is challenging.\footnote{For example, Metropolis-Hastings contains a nondifferentiable operation in the accept/reject step, and HMC additionally requires backpropagating through gradients of the energy function and tuning hyper-parameters.} These complications are avoided when Gibbs sampling of $q_\bphi(\z)$ is possible. The Gibbs transition kernel $q_\bphi(\z_2| \z_1) q_\bphi(\z_1| \z_2')$ of Eq.~\eqref{eq:gibbs_kernel} does not contain an accept/reject step or any hyper-parameters. Assuming that the conditionals can be reparameterized, the gradient of the kernel is approximated efficiently using low-variance reparameterized sampling from each conditional. In this case, the gradient estimator for a single Gibbs update is:
\begin{align}\label{eq:grad_bipartite}
\partial_{\bphi} \E_{q}[f(\z)] &\approx
\E_{q_\bphi(\z')} \bigl[ {\partial_\bphi} \E_{q_\bphi(\z_2| \z_1)q_\bphi(\z_1| \z_2')} [f(\z)] \bigr] \\
&= \E_{q_\bphi(\z')} \E_{\bepsilon_1, \bepsilon_2} \bigl[{\partial_\bphi} f(\z_1, \z_2) \bigr], \nonumber
\end{align}
where {$\z_1(\bepsilon_1, \bphi, \z_2') \sim q_\bphi(\z_1| \z_2')$} and {$\z_2(\bepsilon_2, \bphi, \z_1) \sim q_\bphi(\z_2| \z_1)$}  are reparameterized samples.

Lastly, we note that the reparameterized $\z$ sample has distribution $q_\bphi(\z)$ if $\z'$ is equilibrated. So, if $f$ has its own parameters (e.g., the parameters of the generative model in VAEs), the same sample can be used to compute an unbiased estimation of $\partial_\btheta \E_{q_\bphi(\z)}[f_\btheta(\z)]$ where $\btheta$ denotes the parameters of $f$.

\begin{figure*}[t] 
\vspace{-0.5cm}
    \centering 
\subfloat[KL divergence]{   
\raisebox{0.cm}{\hspace{-0.5cm}\includegraphics[scale=0.27]{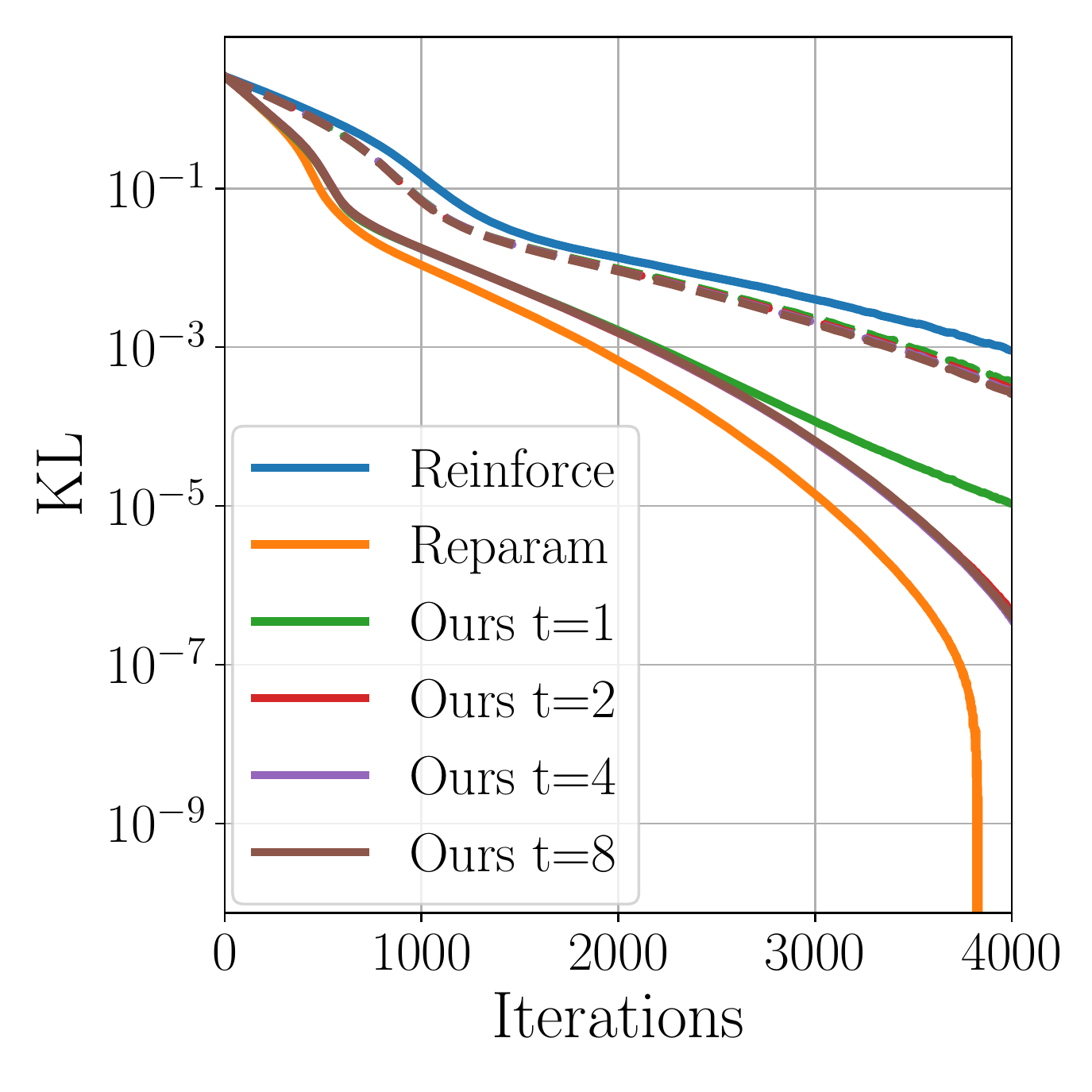}}}
\subfloat[Target distribution]{
\raisebox{-0.03cm}{\hspace{0.4cm}\includegraphics[scale=0.30]{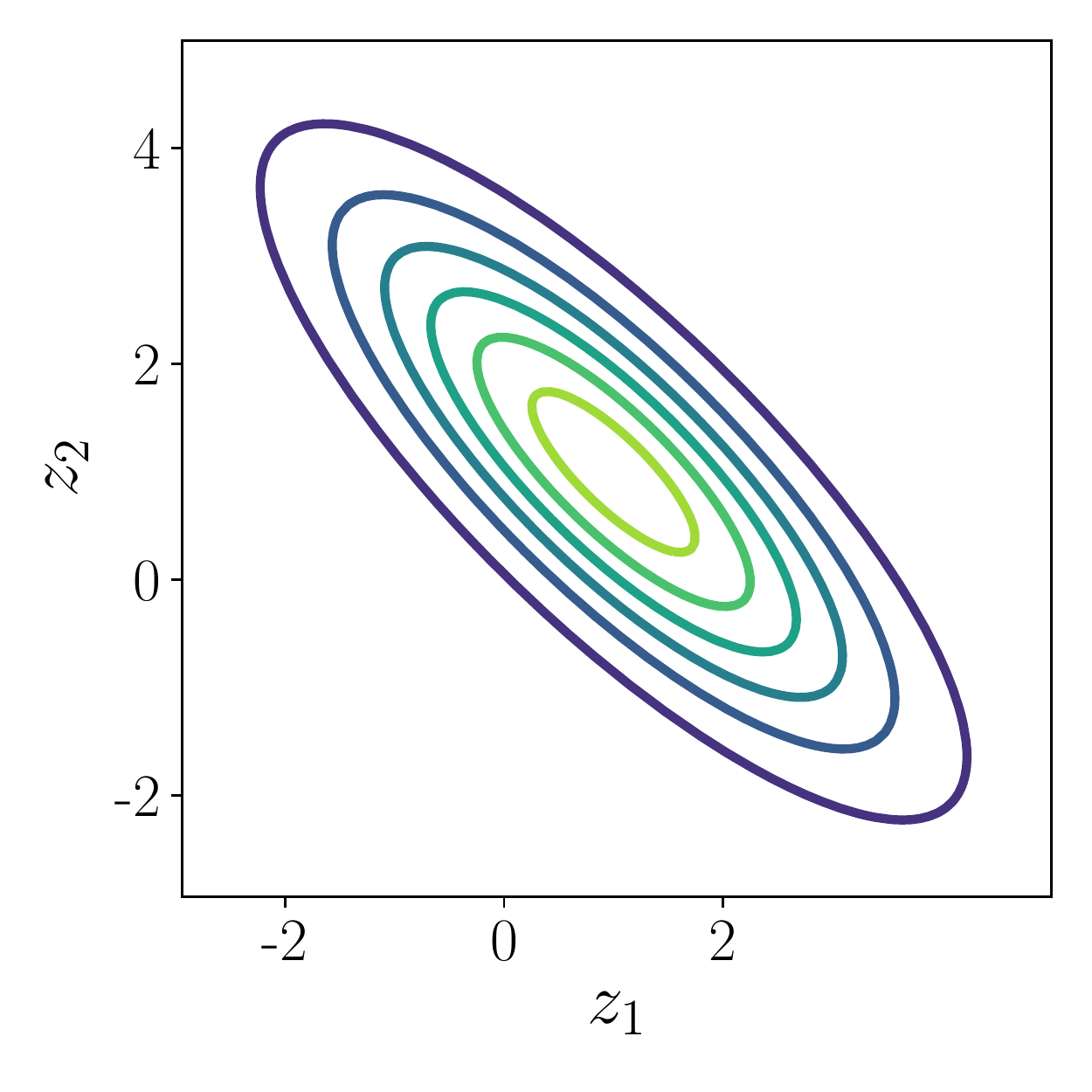}}}
 \subfloat[Gradient $L_2$ error]{   
\raisebox{-0.05cm}{\hspace{0.5cm}\includegraphics[scale=0.31]{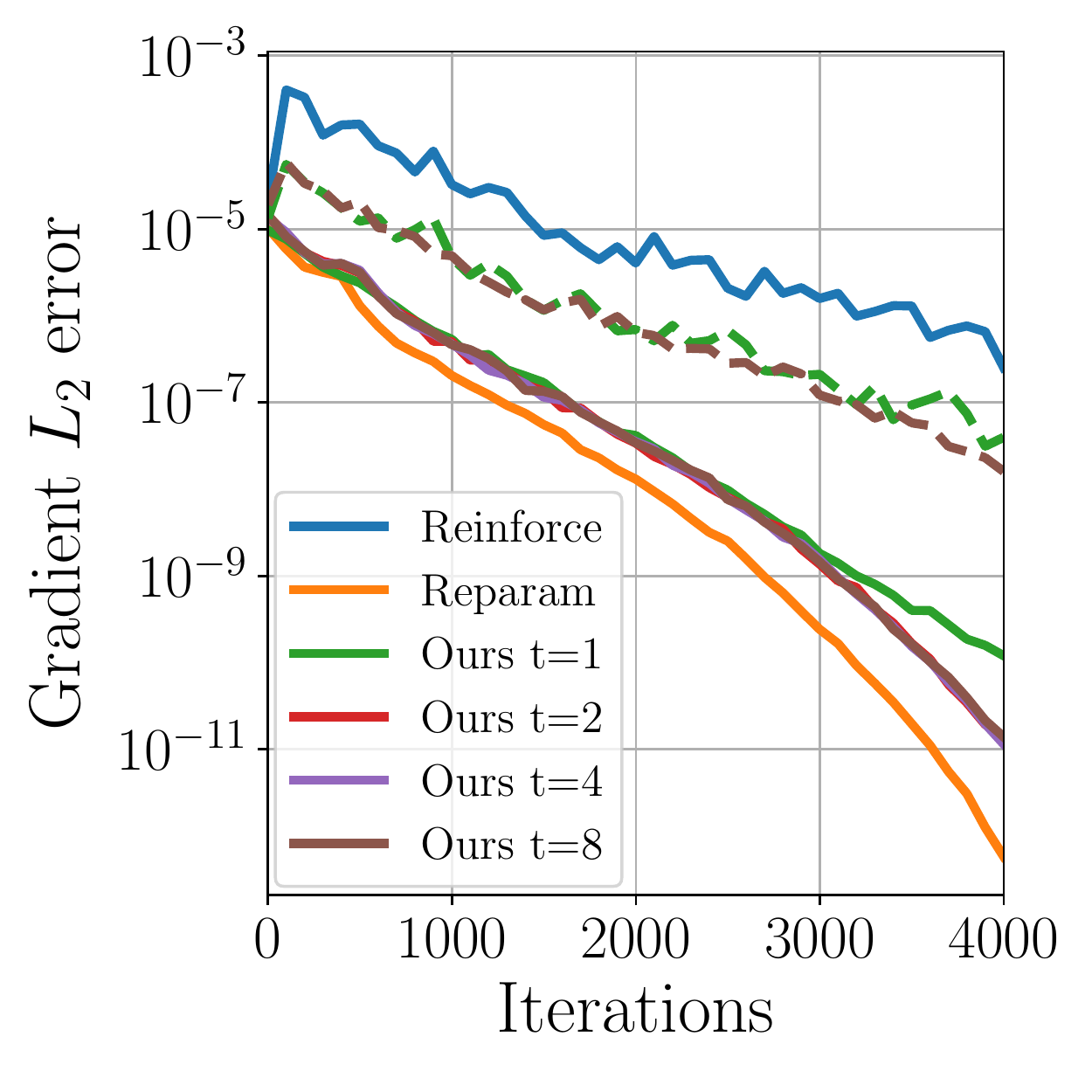}}}
\caption{(a) Our gradient estimator (for various $t$) compared with REINFORCE and reparameterized Gaussian samples for minimizing the KL divergence of a Gaussian to a target distribution (b). Dashed lines correspond to adding the term $I$ of Eq.~\eqref{eq:grad_all} to our gradient estimator.}
    \label{fig:gaussian}
\vspace{-0.3cm}
\end{figure*}
\subsection{Toy Example}\label{sec:toy}

We assess the efficacy of our approximations for a multivariate Gaussian. We consider learning the two-variable Gaussian distribution depicted in Fig.~\ref{fig:gaussian}(b). The target distribution $p(\z)$ has energy function $E(\z)= (\z - \bmu)^T \blambda (\z-\bmu)/2$, where $\pmb{\mu}=[1,1]^T$ and $\blambda = [1.1, 0.9;  0.9, 1.1]$. The approximating distribution $q_\bphi(\z)$ has the same form and we learn the parameters $\bphi$ by minimizing the Kullback-Leibler (KL) divergence ${\rm KL}[q_\bphi(\z) || p(\z)]$. We compare Eq.~\eqref{eq:grad_bipartite} for $t=1, 2, 4, 8$ with reparameterized sampling and with REINFORCE. Fig.~\ref{fig:gaussian}(a) shows the KL divergence during training.
For our method we consider two variants including with and without $I$ in Eq.~\eqref{eq:grad_all}.

Our method significantly outperforms REINFORCE due to lower variance of the gradients as depicted in Fig.~\ref{fig:gaussian}(c). There is little difference between different $t$ until KL divergence becomes $\sim 10^{-4}$. Our method performs worse than reparameterized sampling, which is expected due to the bias introduced by neglecting $I$ in Eq.~\eqref{eq:grad_all}. Including $I$ negatively impacts optimization. The dashed lines of Fig.~\ref{fig:gaussian}(a) show the KL objective when the $I$ term is included. All curves lie atop one another and are noticeably slower to converge. This deterioration is driven by the noisier gradient estimates shown as dashed lines in Fig.~\ref{fig:gaussian}(c). 

Note that this experiment confirms that the reparameterization trick provides low-variance unbiased gradient estimation. However, this trick is not applicable to UGMs in general. In the case of UGMs, our gradient estimator provides gradient estimation with variance properties similar to the reparameterization trick.

\subsection{Learning Boltzmann Machine Posteriors}\label{sec:rbm}
Next, we consider training DVAE\#\#, a DVAE~\citep{rolfe2016discrete, Vahdat2018DVAE++, vahdat2018dvaes} where both approximating posterior and prior are RBM. The objective function of DVAE\#\# is given by Eq.~\eqref{eq:vae}, where $q(\z|\x) \equiv q_{\bphi(\x)}(\z)$ is an amortized RBM with neural-network-generated parameters $\bphi(\x)=\{\b_1(\x), \b_2(\x), \W(\x)\}$. To maximize $\L$, we use the gradient
\be \label{eq:L}
\partial_{\bphi(\x)} \L = \partial_{\bphi(\x)}\E_{q_{\bphi(\x)}(\z)} \left[ \log \frac{p(\x, \z)}{q_{\bphi(\x)}(\z)} \right].
\ee
at each training point $\x$. Although the RBM has a bipartite structure, applying the gradient estimator of Eq.~\eqref{eq:grad_bipartite} is challenging due to binary latent variables. To apply our estimator in Eq.~\eqref{eq:grad_bipartite}, we relax the binary variables to continuous variables using Gumbel-Softmax or piece-wise linear relaxations~\citep{andriyash2018improved}.\footnote{Another option is to use unbiased gradient estimators such as REBAR~\citep{tucker2017rebar} or RELAX~\citep{grathwohl2017backpropagation}. However, with these approaches, the number of $f$ evaluations increases with $t$.} Representing the relaxations of $\z_1 \sim q_{\bphi}(\z_1|\z_2')$ by $\bzeta_1 = \bzeta_1(\bepsilon_1, \z_2', \bphi)$ and $\z_2 \sim q_{\bphi}(\z_2|\z_1)$ by $\bzeta_2 = \bzeta_2(\bepsilon_2, \bzeta_1, \bphi)$, where $\bepsilon_1, \bepsilon_2 \sim \U[0,1]$, we use the following estimator:
\begin{align}
& \E_{q_\bphi(\z_1', \z_2')} \bigl[ {\partial_\bphi} \E_{q_\bphi(\bzeta_1| \z_2')} \E_{q_\bphi(\bzeta_2| \bzeta_1)} [f(\bzeta_1, \bzeta_2)] \bigr]  \nonumber \\
& = \E_{q_\bphi(\z_1', \z_2')} \E_{\bepsilon_1, \bepsilon_2}
   \bigl[ {\partial_\bphi} f(\bzeta_1, \bzeta_2) \bigr], \label{eq:relaxed_gibbs} 
\end{align}
${\partial_\bphi} f(\bzeta_1, \bzeta_2)$ is computed using the reparameterization trick. 

Thus far, we have only accounted for the dependency of the objective $f$ on $\bphi$ through samples $\bzeta$. However, for VAEs, $f$ also depends on $\bphi$ through $\log q_{\bphi(\x)}(\z)$. This dependency can be ignored in VAEs as $\E_{q}[\partial_{\bphi} \log q_\bphi(\z|\x)] = 0$~\citep{roeder2017sticking}, and it can be removed in importance weighted autoencoders using the doubly reparameterized gradient estimation~\citep{tucker2018doubly} (see Appendix~\ref{app:anneal_iw} for more details). 
Note that \citet{roeder2017sticking} and \citet{tucker2018doubly} do not introduce any bias to the gradient and are known to reduce variance.

As $t$ increases and each Gibbs update is relaxed, sampling from the relaxed chain diverges from the exact discrete chain resulting in increasingly biased gradient estimates. Thus, we use $t=1$ in our experiments (see Appendices~\ref{app:toy_example_rbm} and~\ref{app:s_t_expr} for $t>1$). Moreover, in Eq.~\eqref{eq:relaxed_gibbs}, our estimator requires samples from the RBM posterior in the outer expectation. These samples are obtained by running persistent chains for each training datapoint. 

Algorithm~\ref{alg:training_dvae} summarizes training of DVAE\#\#. We represent the Boltzmann prior using $p_\btheta(\z) = \exp\bigl(-E_\btheta(\z) \bigr) / Z_{\btheta}$. The number of Gibbs sweeps for generating $\z'$ in Eq.~\eqref{eq:relaxed_gibbs} is denoted by $s$ and the number of Gibbs sweeps for sampling $\bzeta$ is denoted by $t$. The objective $\mathcal{L}$ is defined so  that automatic differentiation yields the gradient estimator in Eq.~\eqref{eq:relaxed_gibbs} for $\bphi$ and $\partial_{\btheta}f$ is evaluated using relaxed samples for $\btheta$. We use the method introduced by \citet{Vahdat2018DVAE++} Sec.~E to obtain gradients of $\log Z_{\btheta}$.

\vspace{-0.2cm}
\section{Experiments} \label{sec:experiments}
We examine our proposed gradient estimator for training undirected posteriors on three tasks: variational autoencoders, importance weighted autoencoders, and structured prediction models using the binarized MNIST~\citep{salakhutdinov2008dbn} and OMNIGLOT~\citep{lake2015human} datasets. We follow the experimental setup in DVAE\# \citep{vahdat2018dvaes} with minor modifications outlined below.

\subsection{Variational Autoencoders} \label{sec:vae-expt}
We train a VAE of the form $p(\z)p(\x|\z)$, where $p(\z)$ is an RBM and $p(\x|\z)$ is a neural network. $p(\x|\z)$ is represented using a fully-connected neural network having two 200-unit hidden layers, tanh activations, and batch normalization similar to DVAE++~\citep{Vahdat2018DVAE++}, DVAE\#~\citep{vahdat2018dvaes}, and GumBolt~\citep{khoshaman2018gumbolt}.  We compare generative models trained with an undirected posterior (DVAE\#\#) to a directed posterior. 

\setlength{\textfloatsep}{5pt} 
\begin{algorithm}[t]
   \caption{DVAE\#\# with RBM prior and posterior}
   \label{alg:training_dvae}
\begin{algorithmic}
\small  
   \STATE {\bfseries Input:} training sample $\x$,  number of Gibbs sweeps $s$, number of relaxed Gibbs sweeps $t$. 
   \STATE {\bfseries Output:} training objective function $\mathcal{L}_{\x}$
   \STATE $\bphi$ = \texttt{encoder}$(\x)$
   \STATE $\z_{\text{old}}' = $ \texttt{retrieve\_persistent\_states}$(\x)$
   \STATE $\z_{\text{new}}' = $ \texttt{update\_gibbs\_samples}$(\z_{\text{old}}', \bphi$, $s$)
   \STATE $\z_{sg}' = \texttt{stop\_gradient}(\z_{\text{new}}')$  \hfill\algorithmiccomment{$\z_1', \z_2'$ in Eq.~\ref{eq:relaxed_gibbs}}
   \STATE $\bzeta = $
   \texttt{relax\_gibbs\_sample}$(\z_{sg}', \bphi, t)$ \hfill\algorithmiccomment{$\bzeta_1, \bzeta_2$ in Eq.~\ref{eq:relaxed_gibbs}}
   \STATE $\bphi' = $ \texttt{stop\_gradient}$(\bphi)$ \hfill\algorithmiccomment{\citeauthor{roeder2017sticking}}
   \STATE $\mathcal{L}_{\x} = \underbrace{- E_{\btheta}(\bzeta) - \log Z_{\btheta}}_{\text{prior}} + \underbrace{\log p_{\btheta}(\x|\bzeta)}_{\text{likelihood}} + \underbrace{E_{\bphi'}(\bzeta)}_{\text{approx. post}}$
\end{algorithmic}
\end{algorithm}
\setlength{\textfloatsep}{16pt} 

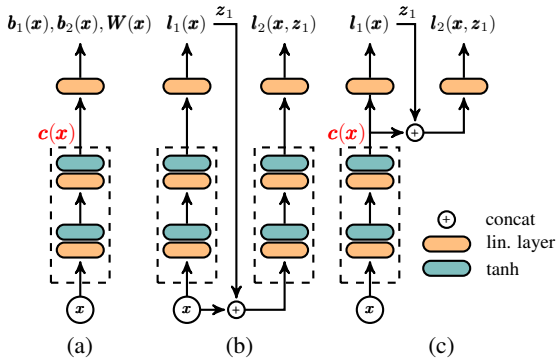
\begin{figure}
    \centering
    \vspace{-0.3cm}
    \hspace{-7.mm}%
  \subfloat[]{
  \begin{tikzpicture}[->,>=stealth',shorten >=1pt,auto,node distance=1.3cm, thick, scale=0.6]
      \tikzstyle{every state}=[fill=white,draw=black,text=black, transform shape, minimum size=0.5cm]
      \tikzstyle{linear} = [draw,scale=0.85, text width=0.5cm,fill=orange!50, rounded rectangle, node distance=0.8cm]
      \tikzstyle{tanh} = [draw,scale=0.85, text width=0.5cm,fill=teal!50, rounded rectangle,node distance=0.28cm]
      \node[state] 			 (x)       {$\x$};
      \node [linear, above=0.5cm of x] (d1) {};
      \node [tanh, above of=d1] (t1) {};
      \node [linear, above of=t1] (d2) {};
      \node [tanh, above of=d2] (t2) {};
      \node [linear, above=0.8cm of t2] (d3) {};
      \node[above=.5cm of d3, scale=0.7]  (z)       {$\b_1(\x), \b_2(\x), \W(\x)$};
      \path (x)         edge              (d1);
      \path (t1)         edge             (d2);
      \path (t2)         edge             (d3);
      \path (d3)         edge             (z);
      
      \draw[dashed] (-0.65,0.6) rectangle (0.65,3.6);
      \node[scale=0.8, red] (c) at (-0.5, 3.9) {$\vc{c}(\x)$};
  \end{tikzpicture}  
  }\hspace{-2.mm}%
  \subfloat[]{
  \begin{tikzpicture}[->,>=stealth',shorten >=1pt,auto, node distance=1.3cm, thick, scale=0.6]
      \tikzstyle{every state}=[fill=white,draw=black,text=black, transform shape,  minimum size=0.5cm];
      \tikzstyle{line} = [draw, -latex', ->];
      \tikzstyle{linear} = [draw,scale=0.85, text width=0.5cm,fill=orange!50, rounded rectangle, node distance=0.8cm]
      \tikzstyle{tanh} = [draw,scale=0.85, text width=0.5cm,fill=teal!50, rounded rectangle,node distance=0.28cm]
      \node[state] 			 (x)       {$\x$};
      \node [linear, above=0.5cm of x] (d1) {};
      \node [tanh, above of=d1] (t1) {};
      \node [linear, above of=t1] (d2) {};
      \node [tanh, above of=d2] (t2) {};
      \node [linear, above=0.8cm of t2] (d3) {};
      \node [linear, right=0.65cm of d1](d4){};
      \node [tanh, above of=d4] (t4) {};
      \node [linear, above of=t4] (d5) {};
      \node [tanh, above of=d5] (t5) {};
      \node [linear, above=0.8cm of t5] (d6) {};
      \node [above=0.5cm of d3, scale=0.7]  (z1)       {$\l_1(\x)$};
      \node [above=0.5cm of d6, scale=0.7]  (z2)       {$\l_2(\x, \z_1)$};
      \node [state,scale=0.4, right=0.6 of x]  (p) {\LARGE $\pmb{+}$};
      \node[scale=0.8] (zz1) at (0.85, 6.6) {\small $\z_1$};
      \path (x)         edge              (d1);
      \path (x)         edge              (p);
      \path (t1)         edge             (d2);
      \path (t2)         edge             (d3);
      \path (d3)         edge             (z1);
      
      \path (t4)         edge             (d5);
      \path (t5)         edge             (d6);
      \path (d6)         edge             (z2);
   
      \path [line] (z1) -| (p);
      \path [line] (p) -| (d4);
      \draw[dashed] (-0.65,0.6) rectangle (0.65,3.6);
      \draw[dashed] (1.5,0.6) rectangle (2.8,3.6);
  \end{tikzpicture}  
  }\hspace{-2mm}%
  \subfloat[]{
  \begin{tikzpicture}[->,>=stealth',shorten >=1pt,auto, node distance=1.5cm, thick, scale=0.6]
      \tikzstyle{every state}=[fill=white,draw=black,text=black, transform shape,  minimum size=0.5cm];
      \tikzstyle{line} = [draw, -latex', ->];
      \tikzstyle{linear} = [draw,scale=0.85, text width=0.5cm,fill=orange!50, rounded rectangle, node distance=0.8cm]
      \tikzstyle{tanh} = [draw,scale=0.85, text width=0.5cm,fill=teal!50, rounded rectangle,node distance=0.28cm]
      \node[state] 			 (x)       {$\x$};
      \node [linear, above=0.5cm of x] (d1) {};
      \node [tanh, above of=d1] (t1) {};
      \node [linear, above of=t1] (d2) {};
      \node [tanh, above of=d2] (t2) {};
      \node [linear, above=0.8cm of t2] (d3) {};
      \node [above=0.5cm of d3, scale=0.7]  (z1)       {$\l_1(\x)$};
      \node [state,scale=0.4, node distance=2.5cm, right of=t2, above=0.3 of t2]  (p) {\LARGE $\pmb{+}$};
      \node[scale=0.8] (zz1) at (0.85, 6.6) {\small $\z_1$};
      
      \node [linear, right=0.6 cm of d3] (d4) {};
      \node [above=0.5cm of d4, scale=0.7]  (z2)       {$\l_2(\x,\z_1)$};
      \node [state,scale=0.4] at (1.7, 2.)  (p1) {\LARGE $\pmb{+}$};
      \node [linear, below=0.1 cm of p1](dd){};
      \node [right=0.1cm of dd, scale=0.8] (tt1) {\footnotesize lin. layer};
      \node [right=0.3cm of p1, scale=0.8] (tt2) {\footnotesize concat};
      \node [tanh, below=0.1 cm of dd](dd2){};
      \node [right=0.1cm of dd2, scale=0.8] (tt2) {\footnotesize tanh};
      \path (x)         edge              (d1);
      \path (t1)         edge             (d2);
      \path (t2)         edge             (d3);
      \path (d3)         edge             (z1);
      \path (d4)         edge             (z2);
      
      \path [line] (z1) -| (p);
      \path [line] (d3) |- (p);
      \path [line] (p) -| (d4);
      
      \draw[dashed] (-0.65,0.6) rectangle (0.65,3.6);
      \node[scale=0.8, red] (c) at (-0.5, 3.9) {$\vc{c}(\x)$};
  \end{tikzpicture}  
  }
  \caption{Neural networks representing $q(\z|\x)$: (a) A 2-layer network predicts the parameters of RBM. (b) The directed posterior used in DVAE\# consists of parallel 2-layer networks to successively predict, $\l$, the logits for each conditional in $q(\z|\x)=\prod_i q_i(\z_i|\x, \z_{<i})$. (c) Our directed posterior differs from DVAE\# and predicts the parameters of each conditional in $q(\z|\x)=\prod_i q_i(\z_i|\vc{c}(\x), \z_{<i})$ using a linear transformation given the shared context feature $\vc{c}(\x)$ and previous $\z$.}   \label{fig:posterior}
\end{figure}

For DVAE\#\#, $q(\z|\x)$ is modeled with a neural network having two tanh hidden layers that predicts the parameters of the RBM (Fig.~\ref{fig:posterior}(a)). Training DVAE\#\# is done using Algorithm~\ref{alg:training_dvae} with $s=10$ and $t=1$ using a piece-wise linear relaxation \citep{andriyash2018improved} for relaxed Gibbs samples. We follow \citep{vahdat2018dvaes} for batch size, learning rate schedule, and KL warm up parameters. During training, samples from the prior are required for computing the gradient of $\log Z_{\btheta}$. This sampling is done using the \citet{QuPA} library that offers population-annealing-based sampling and partition function estimation (using AIS) from within Tensorflow. We also use QuPA for sampling the undirected posteriors and estimating their partition function during evaluation. We explore VAEs with equally-sized RBM prior and posteriors consisting of either 200 (100$+$100) or 400 (200$+$200) latent variables. Test set negative log-likelihoods are computed using 4000 importance-weighted samples.

\textbf{Baselines:} We compare the RBM posteriors with directed posteriors where the posteriors are factored across groups of latent variables $\z_i$ as $q(\z|\x) = \prod_i^{L} q_i(\z_i|\x, \z_{<i})$. Each conditional, $q_i$, is factorial across the components of $\z_i$ and $L$ is the number of hierarchical layers. A fair comparison between directed and undirected posteriors is challenging because they differ in structure and in the number of parameters. We design a baseline so that the number of parameters and the number of nonlinearities is identical for a directed posterior with a single group ($L=1$) and an undirected posterior with no pairwise interactions. This is reasonable as both cases reduce to a mean-field posterior. In Appendix~\ref{app:shared_context}, we present a new structure for directed posteriors with shared context (Fig.~\ref{fig:posterior}(c)) that further improves the posteriors used in DVAE\#~\citep{vahdat2018dvaes} and GumBolt~\citep{khoshaman2018gumbolt} (Fig.~\ref{fig:posterior}(b)).

We examine three recent methods for training VAEs with directed posteriors with shared context. These baselines are i) DVAE\#~\citep{vahdat2018dvaes} that uses power-function distribution to relax the objective function, ii) Concrete relaxation used in GumBolt~\citep{khoshaman2018gumbolt}, and iii) piece-wise linear (PWL) relaxation~\citep{andriyash2018improved}. All models are trained using $L=1$, 2, or 4 hierarchical levels. 

\textbf{Results:} The performance of DVAE\#\# is compared against the baselines in Table~\ref{table:vae_expr}. We make two observations. i) The baselines with $L=1$ are equivalent to a mean-field posterior which is also the special case of an undirected posterior with no pairwise terms. Since PWL and DVAE\#\# use the same continuous relaxation, we can compare PWL $L=1$ and DVAE\#\# to see how introducing pairwise interactions improve the quality of the DVAE models.  We consistently observe $\sim$$0.5$-nat improvement arising from pairwise interactions. ii) DVAE\#\# with undirected posteriors outperforms the directed baselines in most cases indicating that UGMs form more appropriate posteriors.

\textbf{Varying the number of Gibbs steps:} In Appendix~\ref{app:s_t_expr}, we study the effect of changing $s$ and $t$.

\textbf{Using MCMC to sample from the true posterior:} In Appendix~\ref{app:mcmc_expr}, we compare our DVAE\#\#  against MCMC methods that approximately sample from the true posterior.
 
\begin{figure}
    \centering
    \vspace{-0.4cm}
    \subfloat{\hspace{-0.1cm} \includegraphics[scale=0.55, trim={0.4cm, 0, 0, 0}, clip=True]{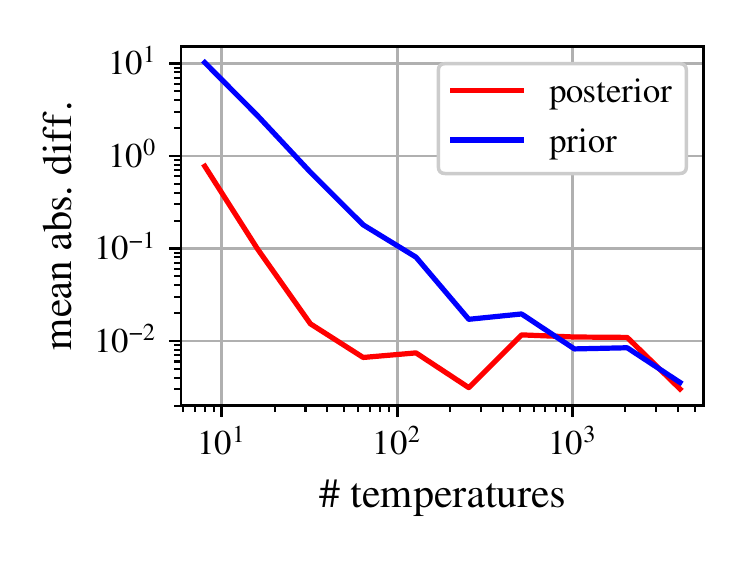}} 
    \subfloat{\hspace{-0.1cm} \includegraphics[scale=0.55, trim={0.4cm, 0, 0, 0}, clip=True]{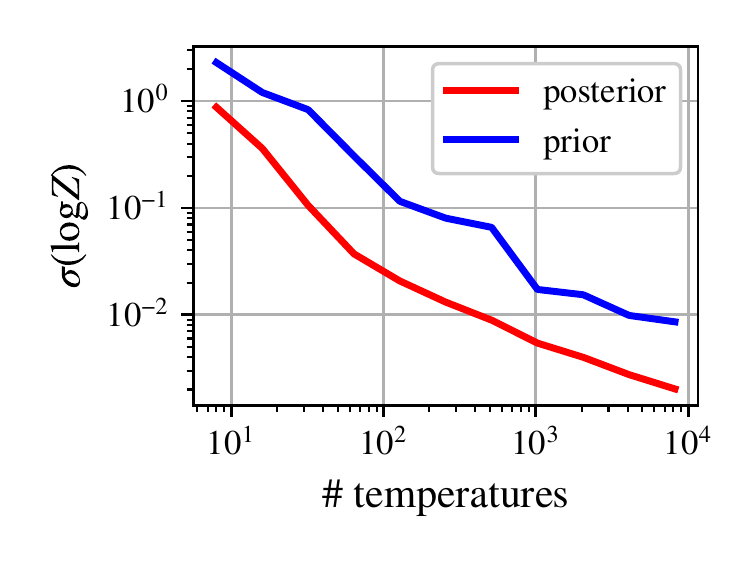}}
    \caption{The difficulty of partition function estimation is visualized using the mean absolute difference between $\log Z$ estimates and its true value in top and the variance of the estimates for different number of temperatures in bottom. The true value is computed using $2^{18}$ temperatures. The number of temperatures required for achieving $\sigma=10^{-2}$ for the RBM posterior is $\sim$10x smaller than the number of temperatures required for the RBM prior.}
    \label{fig:std_logz}
    \vspace{-0.1cm}
\end{figure}

\setlength{\tabcolsep}{3.5pt}
\begin{table*}
\footnotesize
\caption{The performance of DVAE\#\# is compared against directed posteriors, trained with the variational bound. Mean$\pm$standard deviation of the negative log-likelihood for 5 runs are reported. Boldface numbers indicate the best performing models per latent variable size and dataset.
} \label{table:vae_expr}
\centering
\begin{tabular}{|c|c|c|c|c|c||c|c|c|c|}
\hline
& & \multicolumn{4}{c||}{Prior RBM: 100$+$100}  & 
\multicolumn{4}{c|}{Prior RBM: 200$+$200} \\
\hline
& $L$ & DVAE\# & GumBolt & PWL & DVAE\#\# &
DVAE\# & GumBolt & PWL & DVAE\#\# \\ \hline
\multirow{3}{*}{\tiny \begin{turn}{90}\textbf{MNIST}\end{turn}} 
 & 1 & 84.97{\tiny$\pm$0.03} & 84.66{\tiny$\pm$0.04} & 84.62{\tiny$\pm$0.05} & \multirow{3}{*}{\textbf{84.09{\tiny$\pm$0.06}}} & 83.21{\tiny$\pm$0.03} & 83.19{\tiny$\pm$0.05} & 83.22{\tiny$\pm$0.05} & \multirow{3}{*}{\textbf{82.75{\tiny$\pm$0.05}}}\\
 & 2 & 84.96{\tiny$\pm$0.05} & 84.71{\tiny$\pm$0.03} & 84.50{\tiny$\pm$0.07} &  & 83.13{\tiny$\pm$0.04} & 83.04{\tiny$\pm$0.02} & 82.99{\tiny$\pm$0.05} & \\
 & 4 & 84.58{\tiny$\pm$0.02} & 84.39{\tiny$\pm$0.04} & \textbf{84.07{\tiny$\pm$0.04}} &  & 82.93{\tiny$\pm$0.04} & 83.14{\tiny$\pm$0.04} & 82.90{\tiny$\pm$0.03} & \\
\hline
\multirow{3}{*}{\tiny \begin{turn}{90}\textbf{OMNI.}\end{turn}} 
 & 1 & 101.64{\tiny$\pm$0.05} & 101.41{\tiny$\pm$0.06} & 101.24{\tiny$\pm$0.02} & \multirow{3}{*}{\textbf{100.69{\tiny$\pm$0.04}}} & 99.51{\tiny$\pm$0.03} & 99.39{\tiny$\pm$0.04} & 99.32{\tiny$\pm$0.04} & \multirow{3}{*}{\textbf{98.61{\tiny$\pm$0.08}}}\\
 & 2 & 101.75{\tiny$\pm$0.04} & 101.39{\tiny$\pm$0.06} & 101.14{\tiny$\pm$0.05} &  & 99.40{\tiny$\pm$0.05} & 99.12{\tiny$\pm$0.07} & 99.10{\tiny$\pm$0.04} & \\
 & 4 & 101.74{\tiny$\pm$0.07} & 102.04{\tiny$\pm$0.10} & 101.14{\tiny$\pm$0.05} &  & 99.47{\tiny$\pm$0.04} & 99.97{\tiny$\pm$0.02} & 99.30{\tiny$\pm$0.03} & \\
\hline
\end{tabular}
\vspace{0.5cm}
\setlength{\tabcolsep}{3.pt}
\footnotesize
\caption{The performance of DVAE\#\# is compared with directed posteriors with the PWL relaxation trained with the IW bound. Mean$\pm$standard deviation of the negative log-likelihood for 5 runs are reported. Boldface numbers indicate the best performing models per latent space size, dataset, and $K$.
} \label{table:iw_expr}
\centering
\begin{tabular}{|c|c|c|c|c|c||c|c|c|c|}
\hline
& & \multicolumn{4}{c||}{Prior RBM: 100$+$100}  & 
\multicolumn{4}{c|}{Prior RBM: 200$+$200} \\
\hline
& $K$ & PWL {\scriptsize $L=1$} & PWL {\scriptsize $L=2$} & PWL {\scriptsize $L=4$} & DVAE\#\# &
PWL {\scriptsize $L=1$} & PWL {\scriptsize $L=2$} & PWL {\scriptsize $L=4$} & DVAE\#\# \\ \hline
\multirow{3}{*}{\tiny \begin{turn}{90}\textbf{MNIST}\end{turn}} 
 & 1 & 84.60{\tiny$\pm$0.04} & 84.49{\tiny$\pm$0.03} & \textbf{84.04{\tiny$\pm$0.03}} & \textbf{84.06{\tiny$\pm$0.03}} & 83.27{\tiny$\pm$0.06} & 82.99{\tiny$\pm$0.04} & 82.89{\tiny$\pm$0.03} & \textbf{82.76{\tiny$\pm$0.05}}\\
 & 5 & 84.15{\tiny$\pm$0.06} & 83.88{\tiny$\pm$0.02} & \textbf{83.53{\tiny$\pm$0.06}} & \textbf{83.56{\tiny$\pm$0.02}} & 83.24{\tiny$\pm$0.06} & 82.90{\tiny$\pm$0.06} & \textbf{82.65{\tiny$\pm$0.03}} & 82.76{\tiny$\pm$0.06}\\
 & 25 & 83.96{\tiny$\pm$0.05} & 83.70{\tiny$\pm$0.03} & \textbf{83.38{\tiny$\pm$0.02}} & 83.44{\tiny$\pm$0.04} & 83.38{\tiny$\pm$0.04} & 83.05{\tiny$\pm$0.03} & \textbf{82.84{\tiny$\pm$0.02}} & \textbf{82.97{\tiny$\pm$0.11}}\\
\hline
\multirow{3}{*}{\tiny \begin{turn}{90}\textbf{OMNI.}\end{turn}} 
 & 1 & 101.21{\tiny$\pm$0.06} & 101.14{\tiny$\pm$0.07} & 101.12{\tiny$\pm$0.03} & \textbf{100.68{\tiny$\pm$0.05}} & 99.32{\tiny$\pm$0.02} & 99.11{\tiny$\pm$0.02} & 99.28{\tiny$\pm$0.09} & \textbf{98.61{\tiny$\pm$0.06}}\\
 & 5 & 100.72{\tiny$\pm$0.05} & 100.58{\tiny$\pm$0.05} & 100.51{\tiny$\pm$0.04} & \textbf{100.16{\tiny$\pm$0.04}} & 99.19{\tiny$\pm$0.05} & 98.69{\tiny$\pm$0.09} & 98.78{\tiny$\pm$0.02} & \textbf{98.34{\tiny$\pm$0.03}}\\
 & 25 & 100.58{\tiny$\pm$0.01} & 100.48{\tiny$\pm$0.05} & 100.37{\tiny$\pm$0.03} & \textbf{100.02{\tiny$\pm$0.04}} & 99.10{\tiny$\pm$0.03} & 98.70{\tiny$\pm$0.07} & 98.86{\tiny$\pm$0.06} & \textbf{98.34{\tiny$\pm$0.04}}\\

\hline
\end{tabular}
\end{table*}

\subsubsection{Examining Undirected Posteriors} We also examine the posteriors trained with DVAE\#\# on MNIST in terms of the number of modes and the difficulty of sampling and partition function estimation.

\textbf{Multimodality:} To estimate the number of modes, we find a variety of mean field solutions as follows:  Given an RBM we draw samples using \citet{QuPA} and use each sample to initialize the construction of a mean-field approximation. Specifically, we initialize the mean parameter of a factorial Bernoulli distribution to the sample value and then iteratively minimize the KL until we converge to a fixed point describing a mode. This way, the initial population of samples converges to a smaller number of unique modes.

We observe that the majority ($\sim$90\%) of trained posteriors have a single mode. This is in contrast with the prior RBM that typically has 10 to 20 modes. However, we also note that the KL from the converged mean-field distributions to the RBM posteriors is typically in the range $[0.5, 2]$ indicating that, while most RBM posteriors are unimodal, the structure of the mode is not completely captured by a factorial distribution.

\textbf{The difficulty of sampling and partition function estimation:} Fig.~\ref{fig:std_logz} visualizes the deviation of log partition function estimations for an RBM posterior and prior for different numbers of AIS temperatures. As shown in the figure, the number of temperatures required for achieving an acceptable precision (e.g., $\sigma=10^{-2}$) for the RBM posterior is 10 times smaller than the number of temperatures required for the RBM prior. The main reason for this difference is that the RBM posteriors have strong linear biases. When annealing starts from a mean-field distribution containing only the biases, AIS requires fewer interpolating temperatures in order to accurately approximate the partition function. However, RBM priors which do not contain strong linear biases, do not benefit from this.

\subsection{Importance Weighted Autoencoders} \label{sec:iwvae-expt}
Next, we examine the generative model introduced in the previous section but using the importance-weighted (IW) bound. For comparison, we only use the PWL baseline as it achieves the best performance among directed posterior baselines in Table~\ref{table:vae_expr}. For training, we use the same hyperparameters used in the previous section with an additional hyperparameter $K$ representing the number of IW samples.

To sidestep computation of the gradient of the partition function for undirected posteriors in the IW bound, we use the path-wise gradient estimator introduced by \cite{tucker2018doubly}. However, we observe that an annealing scheme in IWAEs (similar to KL-warm up in VAEs) improves the performance of the generative model by preventing the latent variables from turning off. In Appendix~\ref{app:anneal_iw}, we introduce this annealing mechanism for training IWAEs and show how the path-wise gradient can be computed while annealing the objective function.

The experimental results for training DVAE\#\# are compared against directed posteriors in Table~\ref{table:iw_expr}. As can be seen, DVAE\#\# achieves a comparable performance on MNIST, but outperforms the directed posteriors on OMNIGLOT.

\subsection{Structured Output Prediction}\label{sec:sp-expt}
Structured prediction is a form of conditional likelihood estimation~\citep{sohn2015learning} concerned with modeling the distribution of a high-dimensional output given an input. Here, we predict the distribution of the bottom half $\x_2$ of an image given the top half $\x_1$. For this, we follow the IW objective function proposed by~\cite{raiko2014techniques}:
\begin{align}
\E_{\z_{1:K}\sim \prod_i q(\z_i|\x_1)} \bigg[\log \Big( \frac{1}{K} \sum_{i=1}^{K} p(\x_2 | \z_i) \Big) \bigg] + \lambda H(q(\z|\x_1)). \nonumber
\end{align}
We have added an entropy term $H(q(\z|\x_1))$ to prevent the model from over-fitting the training data. This expectation (without the entropy term) is identical to the IW bound in Eq.~\eqref{eq:iw}, where the prior and the approximate posterior are both set to $q(\z|\x_1)$ and it can thus be considered as a lower bound on $\log p(\x_2|\x_1)$. The scalar $\lambda$ is annealed during training from 1 to a small value (e.g., $0.05$).

\begin{figure*}
\begin{minipage}{.48\textwidth}
\captionof{table}{The performance of DVAE\#\# is compared against hierarchical posteriors with the PWL relaxation on the structured prediction problem. 
}
\label{table:struct} 
\centering \hspace{0.1cm}
\setlength{\tabcolsep}{3.pt}
\begin{tabular}{|c|>{\centering\arraybackslash}m{0.3cm}|>{\centering\arraybackslash}m{1.5cm}|>{\centering\arraybackslash}m{1.5cm}|>{\centering\arraybackslash}m{1.5cm}|>{\centering\arraybackslash}m{1.6cm}|}
\hline
& & \multicolumn{4}{c|}{RBM Size: 100$+$100}  \\
\hline
& K &  PWL {\scriptsize $L=1$} & PWL {\scriptsize $L=2$} & PWL {\scriptsize $L=4$} & DVAE\#\#  \\ \hline
\multirow{3}{*}{\tiny \begin{turn}{90}\textbf{MNIST}\end{turn}}
& 1 & 60.82{\tiny$\pm$0.17} & 59.54{\tiny$\pm$0.12} & 60.22{\tiny$\pm$0.12} & \textbf{57.13{\tiny$\pm$0.18}}\\
& 5 & 52.38{\tiny$\pm$0.03} & 52.27{\tiny$\pm$0.16} & 52.67{\tiny$\pm$0.05} & \textbf{49.25{\tiny$\pm$0.07}}\\
& 25 & 48.30{\tiny$\pm$0.08} & 48.41{\tiny$\pm$0.05} & 48.61{\tiny$\pm$0.11} & \textbf{45.75{\tiny$\pm$0.10}}\\
\hline
\multirow{3}{*}{\tiny \begin{turn}{90}\textbf{OMNI.}\end{turn}}
& 1 & 67.06{\tiny$\pm$0.04} & 67.11{\tiny$\pm$0.11} & 67.35{\tiny$\pm$0.08} & \textbf{63.74{\tiny$\pm$0.08}}\\
& 5 & 59.00{\tiny$\pm$0.03} & 59.28{\tiny$\pm$0.08} & 59.34{\tiny$\pm$0.10} & \textbf{57.53{\tiny$\pm$0.08}}\\
& 25 & 54.79{\tiny$\pm$0.04} & 54.83{\tiny$\pm$0.04} & 54.88{\tiny$\pm$0.06} & \textbf{54.32{\tiny$\pm$0.04}}\\
\hline
\end{tabular}
\end{minipage}\hfill
\begin{minipage}{.5\textwidth}
\captionof{table}{Comparison against previously published results on binary VAEs and IWAEs. Our DVAE\#\# outperforms previous models that use similar encoder/decoder.} \label{table:pre_results}
\centering \vspace{-.1cm}
\footnotesize
\setlength{\tabcolsep}{1.0pt}
\begin{tabular}{|l|>{\centering\arraybackslash}m{1.05cm}|>{\centering\arraybackslash}m{1.05cm}|>{\centering\arraybackslash}m{1.05cm}|>{\centering\arraybackslash}m{1.05cm}|}
\hline
& \multicolumn{2}{c|}{\small VAE} & \multicolumn{2}{c|}{\small IWAE}  \\
& \multicolumn{2}{c|}{\tiny $K=1$} & \multicolumn{2}{c|}{\tiny $K\!\in\!\{20, 25, 50\}$}  \\
\hline
&  \tiny MNIST & \tiny OMNI. & \tiny MNIST & \tiny OMNI. \\
\hline
\footnotesize Concrete~(\citeauthor{maddison2016concrete}) & 87.9 & 105.9 & 	85.7 & 106.8 \\ 	
\footnotesize VIMCO~(\citeauthor{mnih2016variational}) & 88.4 & 111.7 & 85.5 & 113.2 \\ 
\footnotesize DVAE++~(\citeauthor{Vahdat2018DVAE++}) & 84.27 & 100.55 & - & - \\ 
\footnotesize GumBolt~(Khoshaman et al.) & 83.28 &	99.83 & \textbf{82.75} &	98.81 \\
\footnotesize GumBolt~\footnotesize(our implement.) & 83.04 & 99.12 & - & - \\
\footnotesize DVAE\#~\footnotesize(\citeauthor{vahdat2018dvaes}) & 83.18 & 99.65 & 82.82 & 98.88 \\
\footnotesize DVAE\#~\footnotesize(our implement.) & 82.93 & 99.40 & - & -\\
\footnotesize DVAE\#\# & \textbf{82.75} & \textbf{98.61} & 82.97 & \textbf{98.34} \\
\hline
\end{tabular}
\end{minipage}
\end{figure*}

Experimental results for the structured prediction problem are reported in Table~\ref{table:struct}. The latent space is limited to 200 binary variables and $q(\z|\x_1)$ and $p(\x_2|\z)$ are modeled by fully-connected networks with one 200-unit hidden layer to limit overfitting. Performance is assessed using the average log-conditional $\log p(\x_2|\x_1)$ measured using 4000 IW samples, and mean$\pm$standard deviation of the negative log-likelihood for five runs are reported. Undirected posteriors outperform the directed models by several nats on MNIST and OMNIGLOT.

\subsection{Comparison against Previously Published Results}
The experiments in this paper are all performed under an identical training framework for all the methods for a fair comparison. However, this requires reimplementing previous work. In order to better situate DVAE\#\# against the state-of-the-art, we compare our performance with previously published results on training binary VAEs that use similar encoder/decoder. Here, for each previous work, we report the best number from the original paper for the latent variable size of 400 with the IW sample size of $K = 1$ and $K \sim 25$. Note that, for the IWAE experiments on DVAE\#\#, $K$ is set to 25, but for a previous work if $K=25$ is not available, we report the closest $K$ in $\{20, 50\}$.

As we can see in Table~\ref{table:pre_results}, our DVAE\#\# model outperforms the previous methods that use similar encoder/decoder. We also observe that our implementation of GumBolt and DVAE\# improves the performance of these models (see Appendix~\ref{app:shared_context} for more details). 

\section{Conclusions and Future Directions}\label{sec:conclusion}
We have introduced a gradient estimator for stochastic optimization with undirected distributions and showed that VAEs and IWAEs with RBM posteriors can outperform similar models having directed posteriors. These encouraging results for UGMs over discrete variables suggest a number of promising research directions.

\textbf{Exponential Family Harmoniums:} 
The methods we have outlined also apply to UGMs defined over continuous variables (as in Fig~\ref{fig:gaussian}). Exponential Family Harmoniums (EFHs)~\citep{welling2005exponential} generalize RBMs and are composed of two disjoint groups of exponential-family random variables.
EFHs retain the factorial structure of the conditionals $q(\z_1 | \z_2)$ and $q(\z_2 | \z_1)$ so that we can easily backpropagate through Gibbs updates. However, the exponential-family generalization allows for the mixing of different latent variable types and distributions.

\textbf{Auxiliary Random Variables:} An appealing property of RBMs (and EFHs) is that either group of variables can be analytically marginalized out. This property can be exploited to form more expressive approximate posteriors. Consider a generative model $p(\z, \x) = p(\z)p(\x|\z)$. To approximate the posterior we augment the latent space of $\z$ with auxiliary variables $\h$, and form a UGM over the joint space. In this case, the marginal approximate posterior $q(\z|\x) = \sum_{h} q(\z, \h|\x)$ has more expressive power. Sampling from $q(\z|\x)$ can be done by sampling from the joint $q(\z, \h|\x)$ and our gradient estimator can be used for training the parameters of the UGM. The objective function of VAE can be optimized easily as the log-marginal $\log q(\z|\x)$ has an analytic expression up to the normalization constant given a sample from the posterior.

\textbf{Combining DGMs and UGMs:} VAEs trained with DGM posteriors $q(\z) = \prod_i q_i(\z_i|\z_{<i})$ have shown promising results. However, each factor $q_i(\z_i|\z_{<i})$ in DGMs is typically assumed to be a product of independent distributions. We can build more powerful DGMs by modeling each $q_i(\z_i|\z_{<i})$ using a UGM. 

\textbf{Sampling and Computing Partition Function:} A challenge when using continuous-variable UGMs as posteriors is the requirement for sampling and partition function estimation during evaluation if the test data log-likelihood is estimated using the IW bound. However, we note that approximate posteriors are only required for training VAEs. At evaluation, we can sidestep sampling and partition function estimation challenges using techniques such as AIS~\citep{wu2016quantitative} starting from the prior distribution when the latent variables are continuous.

{\small
\bibliography{generative.bib}
\bibliographystyle{icml2020}}

\clearpage
\appendix
\section{Toy Example: RBM Posterior} \label{app:toy_example_rbm}
\begin{figure}[t]    
    \centering 
    \hspace{-1.7cm}
    \includegraphics[scale=0.35]{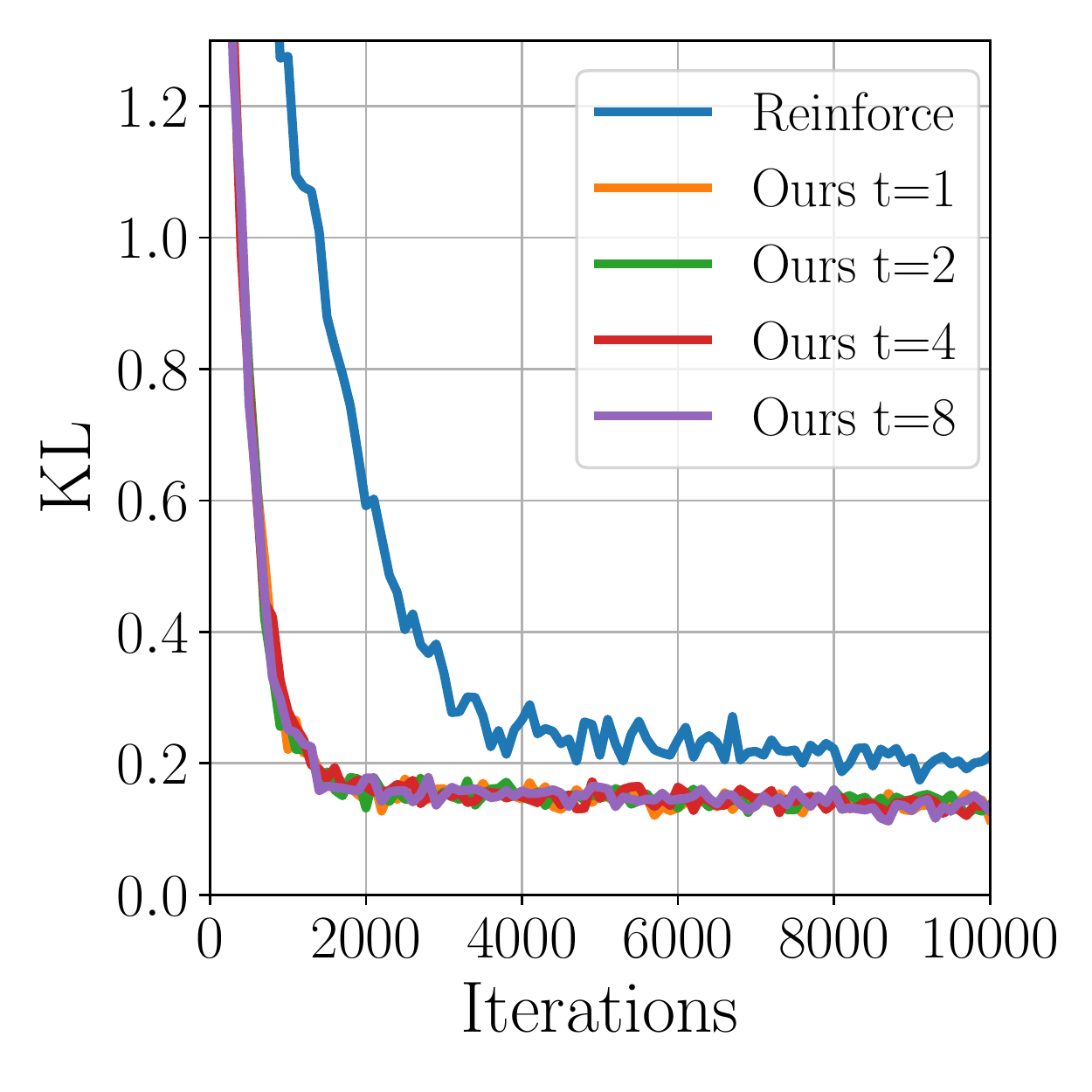}
\caption{Optimization of KL divergence from an RBM to a mixture of Bernoulli distributions using the proposed gradient estimator. Increasing the number of MCMC updates $t$ does not improve the gradient estimation.}
    \label{fig:mixture}
\end{figure}

In this section, we examine the dependence of the gradient estimator the number of MCMC updates $t$ on a toy example. Here, a 4+4 bit RBM is learned to minimize the KL divergence to a mixture of Bernoulli distributions:
\begin{align}
\L & = \E_{q_\bphi(\z)} \Bigl[ \log \frac{q_\bphi(\z)}{p_{\rm mix}(\z)} \Bigr] \\
p_{\rm mix}(\z) &= \sum_{i=1}^C \alpha_i e^{\sum_a \left(  \nu_i^a z^a - {\rm softplus}(\nu_i^a) \right)}.
\end{align}
We choose $C=3$ mixture components with weights $\alpha_i$ chosen uniformly and Bernoulli components having variance $0.09$. The training of parameters $\bphi$ is done with the Adam optimizer~\citep{kingma2014adam}, using learning rate $0.003$ for 10000 iterations. We show the results for several values of $t$ in Fig~\ref{fig:mixture}. In all cases the optimization locates the global minimizer. We note that the effectiveness of optimization does not depend on $t$.

\section{Examining the Shared Context} \label{app:shared_context}
For the directed posterior baselines, we initially experimented with the structure introduced in DVAE\#\footnote{These models are also used in DVAE++ and GumBolt.} in which each factor $q(\z_i|\x, \z_{<i})$ is represented using two tanh nonlinearities as shown in Fig.~\ref{fig:posterior}(b). However, initial experiments indicated that when the number of latent variables is held constant, increasing the number of subgroups, $L$, does not improve the generative performance.

In undirected posteriors, the parameters of the posterior for each training example are predicted using a single shared context feature ($\vc{c}(\x)$ in Fig.~\ref{fig:posterior}(a)). However, in DVAE\#, parallel neural networks generate the parameters of the posterior for each level of the hierarchy. Inspired by this observation, we define a new structure for the directed posteriors by introducing a shared (200 dimensional) context feature that is computed using two tanh nonlinearities. This shared feature $\vc{c}(\x)$ is fed to the subsequent conditional each represented with a linear layer, i.e., $q(\z|\x) = \prod_{i=1}^L q(\z_i|\vc{c}(\x), \z_{<i})$ (see Fig.~\ref{fig:posterior}(c)).

In Table~\ref{table:context}, we compare the original structure used in DVAE\# to the structure with a shared context. In almost all cases the shared context feature improves the performance of the original structure. Moreover, increasing the number of hierarchical layers often improves the performance of the new structure. 

Note that in Table~\ref{table:context}, we report the average negative log-likelihood measured on the test set for 5 runs. In the $L=1$ case, both structures are identical and they achieve statistically similar performance.

\begin{table*}[th]
\small
\caption{ The performance of structure with a shared context feature (Fig.~\ref{fig:posterior}(c)) is compared against the original structure used in DVAE\# (Fig.~\ref{fig:posterior}(b)). The shared context feature often improves the performance. 
} \label{table:context}
\centering
\setlength{\tabcolsep}{2.pt}
\begin{tabular}{|c|c|c|>{\centering\arraybackslash}m{1.0cm}|>{\centering\arraybackslash}m{1.0cm}||>{\centering\arraybackslash}m{1.0cm}|>{\centering\arraybackslash}m{1.0cm}||>{\centering\arraybackslash}m{1.0cm}|>{\centering\arraybackslash}m{1.0cm}||>{\centering\arraybackslash}m{1.0cm}|>{\centering\arraybackslash}m{1.0cm}||>{\centering\arraybackslash}m{1.0cm}|>{\centering\arraybackslash}m{1.0cm}||>{\centering\arraybackslash}m{1.0cm}|>{\centering\arraybackslash}m{1.0cm}|}
\hline
\multicolumn{3}{|c|}{} & \multicolumn{6}{c||}{Prior RBM Size: 100$+$100} & \multicolumn{6}{c|}{Prior RBM Size: 200$+$200}  \\
\hline
\multicolumn{3}{|c|}{} & \multicolumn{2}{c||}{DVAE\#} & \multicolumn{2}{c||}{GumBolt} & \multicolumn{2}{c||}{PWL} & \multicolumn{2}{c||}{DVAE\#} & \multicolumn{2}{c||}{GumBolt} & \multicolumn{2}{c|}{PWL} \\ \hline
\multicolumn{3}{|c|}{context}  & \ding{55} & \checkmark & \ding{55} & \checkmark & \ding{55} & \checkmark & \ding{55} & \checkmark & \ding{55} & \checkmark & \ding{55} & \checkmark \\ \hline
\multirow{3}{*}{\tiny\begin{turn}{90}\textbf{MNIST}\end{turn}} & \multirow{3}{*}{\tiny\begin{turn}{90}\#layers\end{turn}} 
 & 1 & \textbf{84.95} & \textbf{84.97} & \textbf{84.65} & \textbf{84.66} & \textbf{84.57} & \textbf{84.62} & \textbf{83.26} & \textbf{83.21} & \textbf{83.21} & \textbf{83.19} & \textbf{83.23} & \textbf{83.22}\\
 & & 2 & \textbf{84.81} & 84.96 & \textbf{84.75} & \textbf{84.71} & 84.70 & \textbf{84.50} & 83.26 & \textbf{83.13} & 83.24 & \textbf{83.04} & 83.38 & \textbf{82.99}\\
 & & 4 & \textbf{84.61} & \textbf{84.58} & 84.90 & \textbf{84.39} & 85.00 & \textbf{84.07} & 83.13 & \textbf{82.93} & 83.69 & \textbf{83.14} & 83.85 & \textbf{82.90}\\
\hline
\multirow{3}{*}{\tiny \begin{turn}{90}\textbf{OMNI.}\end{turn}} & \multirow{3}{*}{\tiny \begin{turn}{90}\#layers\end{turn}} 
 & 1 & \textbf{101.69} & \textbf{101.64} & \textbf{101.41} & \textbf{101.41} & \textbf{101.21} & \textbf{101.24} & \textbf{99.53} & \textbf{99.51} & \textbf{99.39} & \textbf{99.39} & \textbf{99.37} & \textbf{99.32}\\
 & & 2 & 101.84 & \textbf{101.75} & 102.45 & \textbf{101.39} & 101.64 & \textbf{101.14} & 99.66 & \textbf{99.40} & 100.52 & \textbf{99.12} & 100.07 & \textbf{99.10}\\
 & & 4 & 101.93 & \textbf{101.74} & 102.76 & \textbf{102.04} & 101.97 & \textbf{101.14} & 99.63 & \textbf{99.47} & 100.99 & \textbf{99.97} & 100.45 & \textbf{99.30}\\
\hline
\end{tabular}
\end{table*}

\section{Annealing the Importance Weighted Bound} \label{app:anneal_iw}
KL annealing~\citep{sonderby2016ladder} is used to prevent latent variables from turning off in VAE training. A scalar $\lambda$ is introduced which weights KL contribution to the ELBO and $\lambda$ is annealed from zero to one during training:
\be \label{eq:kl_annealing}
\L = \E_{q(\z|\x)} [\log p(\x|\z)] - \lambda \KL\bigl(q(\z|\x) || p(\z)\bigr).
\ee
Since the KL contribution is small early in training, the VAE model initially optimizes the reconstruction term which prevents the approximate posterior from matching to the prior.

Here, we apply the same approach to the importance weighted bound by rewriting the IW bound by:
\be \label{eq:iw_annealing}
\L_{IW} = \E_{\z_{1:K}} \left[ \log \left( \frac{1}{K} \sum_{i=1}^{K} \frac{p(\z_i)^\lambda p(\x|\z_i)}{q(\z_i|\x)^\lambda} \right) \right],
\ee
where $\z_{1:K}\sim \prod_{i=1}^K q(\z_i|\x)$. Similar to the KL annealing idea, we anneal the scalar $\lambda$ during training from zero to one. When $\lambda$ is small, the IW bound emphasizes reconstruction of the data which inhibits the latent variables from turning off. When $K=1$, Eq.~\eqref{eq:iw_annealing} reduces to Eq.~\eqref{eq:kl_annealing}.

The gradient of Eq.~\eqref{eq:iw_annealing} with respect to $\bphi$, the parameters of $q(\z|\x)$ and $\btheta$, the parameters of the generative model is:
\be \label{eq:grad_iw}
\partial_{\bphi, \btheta} \L_{IW} = \E_{\bepsilon_{1:K}} \left[ \sum_{i=1}^{K} \frac{w_i}{\sum_j w_j} \partial_{\bphi, \btheta} \log w_i \right],
\ee
where $w_i = p_{\btheta}(\z_i)^\lambda p_{\btheta}(\x|\z_i) / q_{\bphi}(\z_i|\x)^\lambda$.
The gradient $\partial_{\bphi} \log w_i$ is evaluated as:
\be \label{eq:grad_log_w}
\partial_{\bphi} \log w_i = - \lambda \partial_\bphi \log q_{\bphi}(\z_i|\x) + (\partial_{\z_i}\!\! \log w_i)(\partial_\bphi \z_i).
\ee
The second term in the right hand side of Eq.~\eqref{eq:grad_log_w} is the path-wise gradient that can be computed easily using our biased gradient estimator. This term does not have any dependence on the partition function $Z_{\bphi}$ since
\be
{\partial_{\z_i}} \! \log w_i = {\partial_{\z_i}} \! \left( \log p_{\btheta}(\z_i)^\lambda p_{\btheta}(\x|\z_i) - \lambda \log q_{\bphi}(\z_i|\x) \right) \nonumber
\ee
and ${\partial_{\z_i}}\log q_{\bphi}(\z_i|\x) = -{\partial_{\z_i}} E_{\bphi}(\z_i)$ depends only on the energy function. 

However, the first term in the right hand side of Eq.~\eqref{eq:grad_log_w} does contain ${\partial_\bphi} Z_{\bphi}$ which can be high-variance. We apply the doubly reparameterized method proposed by \citep{tucker2018doubly} to remove this term. Following a derivation similar to \citep{tucker2018doubly}, it is easy to show that:
\begin{equation} 
\partial_{\bphi} \L_{IW} = \E_{\bepsilon_{1:K}} \! \left[ 
\sum_{i=1}^{K} \left( \lambda \tilde{w_i}^2 + (1 - \lambda) \tilde{w_i} \right) \! (\partial_{\z_i} \! \log w_i) (\partial_\bphi \z_i) \right], \nonumber
\end{equation}
where $\tilde{w_i}=w_i / \sum_j w_j$.

\section{The Effect of Gibbs Chain Lengths $\pmb{s}$ and $\pmb{t}$}\label{app:s_t_expr}
In Table.~\ref{table:s_t_expr}, we study the effect of changing the number of Gibbs sampling steps on the final performance of DVAE\#\#. We vary the number of discrete Gibbs steps $s$ and the number of relaxed Gibbs steps $t$ in training the DVAE\#\# model on MNIST with the variational bound for an RBM size of $100+100$. We observe that the final performance does not change for $s \ge 5$, however, increasing $t$ decreases the final log likelihood. This degradation with increasing $t$ likely arises as sampling from the relaxed chain diverges from the exact discrete chain resulting in increasingly biased gradient estimates. 

\begin{table}
\footnotesize
\centering
{
\caption{The performance of DVAE\#\# trained on MNIST with the variational bound is compared for varying numbers of discrete Gibbs steps $s$ and relaxed Gibbs steps $t$. The mean$\pm$standard deviation of the negative log-likelihood over 4 runs is reported. Final performance does not change for $s \ge 5$, but increasing $t$ degrades performance.
} \label{table:s_t_expr}
\centering
\vspace{0.2cm}
\begin{tabular}{|c|c|c|c|}
\hline
\multicolumn{4}{|c|}{RBM Size: 100$+$100}  \\
\hline
      & $s=1$ & $s=5$ & $s=10$\\ \hline
$t=1$ & \textbf{84.11{\tiny$\pm$0.04}}      &   \textbf{84.06{\tiny$\pm$0.04}}    &   \textbf{84.09{\tiny$\pm$0.03}}    \\ \hline
$t=2$ & 84.20{\tiny$\pm$0.05}      &   \textbf{84.12{\tiny$\pm$0.04}}    &   \textbf{84.12{\tiny$\pm$0.06}}    \\ \hline
$t=4$ & 84.46{\tiny$\pm$0.04}      &   84.32{\tiny$\pm$0.05}    &   84.34{\tiny$\pm$0.02}    \\ \hline
\end{tabular}}
\end{table}

\section{MCMC Sampling from True Posterior}\label{app:mcmc_expr}
A natural question to ask is how our method compares against MCMC approaches, which approximately sample from the true posterior, instead of sampling from an amortized undirected approximate posterior. In this section, we provide a small experiment to compare these two approaches.

Training a generative model using MCMC is similar to training a VAE, with the main difference that, instead of using an encoder to sample from an approximate posterior, we use MCMC to approximately sample from the true posterior. Given the samples from true posterior, we use the following objective to train the parameters of the generative model ($\btheta$):
\be \label{eq:mcmc_obj}
\max_{\btheta} \E_{\z\sim p_{\btheta}(\z|\x)} \left[ \log p_\btheta(\x, \z)\right].
\ee
There are several pitfalls in using MCMC to sample from the true posterior in generative models. First, we mainly consider binary latent variables in this paper. Thus, gradient-based MCMC techniques such as Hamiltonian Monte Carlo are not available. To overcome this, we sample one bit at a time from the true posterior, and in each sweep, we iterate over all the bits. Second, MCMC may take many iterations to reach equilibrium. To address this, we use persistent sampling for each data point in the training set. We start the MCMC chains from the state of the chains from the previous epoch, and we use 10 sweeps before each parameter update. Finally, in the absence of encoders. computing the log-likelihood of the MCMC approaches on the test dataset is not trivial. The best known approach is AIS~\cite{wu2016quantitative} which is challenging to implement. To address this, we limit the latent space to only 16 bits, and we compute log-likelihood exactly by enumerating all the states. 

In Table~\ref{table:mcmc_expr}, we report the negative log-likelihood on the training and test datasets and training time per iteration. We make two observations. First, the training log-likelihood for both RBM posteriors and the true posteriors are not significantly different. However, the test log-likelihood is inferior to the MCMC baseline. When we examine training the MCMC baseline, we notice that the training dynamic for the MCMC method is very different than DVAE\#\#. MCMC training is sensitive to the initialization and is prone to overfitting. We hypothesize that MCMC methods require their own special treatment (beyond the scope of this paper), making a direct comparison against VAE models unfair.

Second, we note that training using MCMC sampling from the posterior is $\sim$5$\times$ slower than training using the RBM approximate posteriors. This is mostly because, in the RBM approximate posteriors, each conditional in the transition kernel is a linear function. However, in the MCMC sampling from the true posterior, we require computing the likelihood $p(\x|\z)$ in each sampling step, which is computationally expensive as $p(\x|\z)$ is implemented by a neural network. We expect MCMC approaches to be even slower as the generative model becomes deeper and more complex while the computational complexity of sampling from the RBM approximate posteriors does not depend on the generative model.
\begin{table}
\footnotesize
\centering
{
\caption{The performance of DVAE\#\# trained on MNIST with the variational bound is compared against using MCMC to sample from the true posterior. Here, we limit the latent space to 16 bits. The negative log-likelihood on the training and test dataset and training time per iteration in milliseconds are reported.
} \label{table:mcmc_expr}
\centering
\vspace{0.2cm}
\begin{tabular}{|c|c|c|}
\hline
\multicolumn{3}{|c|}{Prior Size: 16 bits}  \\
\hline
      & DVAE\#\# & MCMC \\
      & RBM approx. posterior & True posterior \\ \hline
train neg. LL & \textbf{105.7{\tiny$\pm$0.3}}      &   \textbf{106.7{\tiny$\pm$1.3}}      \\ \hline
test neg. LL & \textbf{110.9{\tiny$\pm$0.3}}      &   113.4{\tiny$\pm$1.5}      \\ \hline
Time (ms) & \textbf{22}      &   103       \\ \hline
\end{tabular}}
\end{table}

\end{document}